# A generative-informed neuro-symbolic framework for syntactic ambiguity resolution: Evidence from Arabic DPs

**Mohammed Damom, Hajjah University**
**Mohammed Q. Shormani, Ibb University**
shormani@ibbuniv.edu.ye/https://orcid.org/0000-0002-0138-4793
**Muneef Y. Alshawsh, Ibb University**
**Ashraf A. Naji, Ibb University**
**Mustafa Ali Alhamzi, Hajjah University**
**Fawwaz An-Nashef, Sana'a University**
**Jameel Ahmed Elayah, Ibb University**
**Noman AL-Sayadi, Ibb University**
**(October 2026)**

## Abstract

Syntactic ambiguity poses a persistent challenge for Arabic NLP, particularly in morphologically rich nominal constructions where multiple structu6ral interpretations may be compatible with the same surface sequence. This study proposes a generatively informed neuro-symbolic framework for resolving structural ambiguity in Modern Standard Arabic (MSA) DPs. The framework integrates generative syntactic notions with AraBERT by representing ambiguity as a candidate-based decision task in which linguistically motivated alternatives are explicitly constructed and evaluated through candidate-conditioned input representations. Findings indicate that the model achieved 96.88% accuracy, 95.92% macro-F1, 96.83% weighted F1, and 93.94% binary F1 on the unseen evaluation set. Class-level analysis revealed asymmetric performance, with recall of 99.71% for High/VP Attachment (N1) and 89.26% for Low/NP/Embedded Attachment (N2), indicating greater difficulty in recovering the embedded interpretation. The study concludes that formal syntactic representations can be operationalized within Transformer-based NLP as an explicit interface between linguistic structure and contextual neural modeling, providing a controlled and interpretable approach to Arabic syntactic ambiguity resolution and beyond.

**Keywords:** Generative linguistics, neuro-symbolic mechanism, *Search Space, Merge*, Arabic NLP, syntactic ambiguity resolution, Arabic DPs, AraBERT, candidate-based disambiguation

## 1. Introduction

The mainstream line of deep learning (DL) research frequently confronts fundamental questions regarding syntactic phenomena like structural ambiguity that present well-documented processing challenges even for native human speakers (Shormani, 2014a). Central to this inquiry is whether Transformer-based models (TBMs) like BERT, AraBERT (Antoun et al., 2020), and MARBERT (Abdel-Mageed & Elmadany, 2021; Abdel-Mageed et al., 2022) truly comprehend natural language ambiguity. If they do, critical questions remain about the extent to which their contextual representations capture structural variations, the exact computational mechanisms driving their disambiguation capacity, and whether integrating formal generative syntax assumptions and structured inputs can enhance their performance.

Ambiguity is an inherent property of natural language (Chomsky, 2002), spanning phonological, lexical, and syntactic dimensions. Syntactic ambiguity has long constituted a central problem in linguistic theory because a single surface string may correspond to more than one underlying syntactic interpretation (Chomsky, 1965; Frazier & Fodor, 1978). This phenomenon is particularly significant within generative approaches to syntax, where the relationship between overt linguistic form and covert structural representations is a primary theoretical concern. From this perspective, structural ambiguity is not merely an artifact of lexical polysemy or pragmatic context; rather, it emerges when a sequence of surface constituents permits more than one syntactic organization, attachment relation, or hierarchical derivation. Thus, examining syntactic ambiguity provides an essential empirical benchmark for evaluating how overt form maps onto alternative structural interpretations in both human sentence processing and artificial neural networks (Manning et al., 2020).

In this study, we investigate syntactic ambiguity within Arabic Determiner Phrases (DPs), focusing specifically on Construct States (CSs/*iḍāfa*), nonCSs and coordinated DPs. Arabic DP constructions exhibit rich morphosyntactic agreements like φ-features, and definiteness constraints, that interact directly with attachment height, yielding well-defined structural ambiguities (Shormani, 2014a). This investigation addresses an unexplored empirical domain, as no previous work has targeted attachment ambiguity resolution in Arabic CSs by combining Transformer-based contextual encoders with formal syntactic candidate generation. Additionally, the structural precision of Arabic CS attachment offers a controlled testing ground to determine whether neural language models learn genuine hierarchical syntactic relations or rely on shallow lexical association biases (Linzen et al., 2016; Gulordava et al., 2018). Ultimately, this work establishes a framework for incorporating formal generative linguistics into modern machine learning (ML) architectures, demonstrating how symbolic generative constraints and neural contextual representations can be synthesized into a unified model.

In this study, we investigate a compelling topic focusing strictly on syntactic ambiguity in Arabic DPs, including CSs, non-NonCSs, and coordinated DPs. We conduct this study for several reasons: i) to the best of our knowledge, no previous study has computationally modeled this specific structural contrast by bridging the symbolic operations of generative syntax, specifically *Merge* (building hierarchical structures), *Agree* (enforcing feature-matching constraints like φ-features, Case), and the definition of a constrained structural search space, with the contextual attention mechanisms of neural Transformer models, ii) existing Arabic NLP benchmarks lack fine-grained, human-annotated ground-truth corpora dedicated to structural DP attachment ambiguity. We bridge this gap by establishing a verified dataset of 6696 human-ground-truth instances (4881 High/VP attachment cases and 1815 Low/NP attachment cases) systematically filtered from an initial pool of 44637 sentence-level candidates, and iii) our finetuned AraBERT model demonstrates that neural multi-head attention can effectively learn candidate-conditioned structural preferences, achieving notable performance across complex ambiguity configurations while highlighting remaining challenges in resolved vs. unresolved feature-agreement contexts. Thus, this study opens new horizons for investigation within Arabic computational syntax, establishing a promising neuro-symbolic framework that demonstrates how formal generative primitives and deep neural representations can mutually inform one another.

The remainder of this article is structured as follows. Section 2 outlines the theoretical foundations, reviews related work in Arabic syntactic parsing and structural ambiguity resolution. Section 3 spells out the study methodology, candidate-based representation, dataset construction, and model training procedures. Section 4 presents the experimental results and fine-grained error analysis. and Section 5 discusses these results, and Section 6 concludes the article with main implications, limitations and directions for future research.

## 2. Conceptual Framework

### 2.1. Generative syntax

Generative syntax has developed around a fundamental question: What computational system underlies human syntactic knowledge, and how does this system generate and interpret structurally organized linguistic expressions? From Chomsky's *Syntactic Structures* (1957) and *Aspects of the Theory of Syntax* (1965) through Principles and Parameters (P&P) and Minimalism, generative syntax has progressively shifted towards a more explicit and restrictive account of syntactic structure and computation. For this study, this development is important because syntactic ambiguity concerns how a single surface expression can correspond to alternative structural representations. The early transformational-generative tradition established that linguistic knowledge cannot be reduced to linear word order. Subsequent developments, particularly Government and Binding (Chomsky, 1981), articulated syntactic structure through interacting principles governing such phenomena as constituent structure, Case, binding, movement, and thematic relations. This perspective is directly relevant to ambiguity: when the same sequence of words permits more than one interpretation, the alternatives must be distinguished structurally rather than merely lexically or linearly.

The relevance of this approach is specifically clear in Arabic, as it provides a rich empirical domain for investigating the interaction of constituent structure, Case, agreement, movement, word order, morphology, and interpretation. Research on Standard Arabic has argued, for example, that surface order alone is insufficient to determine clause structure, with evidence from Case, agreement, movement, binding, negation, modality, aspect, discourse interpretation (Shormani, 2017), and ambiguity contributing to structural analysis (Shormani, 2014a). Such work illustrates a methodological principle central to this study, precisely syntactic structure must be established through converging syntactic and interpretive evidence rather than surface order alone.

Minimalism marked a further development by asking how much of the syntactic machinery apparatus could be reduced while retaining an adequate account of syntactic computation. Operations such as *Merge* (E/I), *Agree*, *Feature Valuation*, and *Spell-out* became central to explaining how hierarchical structures are constructed and interpreted at the interfaces. The most important notion of Minimalism is perhaps the *Strong Minimalist Thesis* (SMT) which states that "*Language is an optimal solution to legibility conditions"* Chomsky (2000: 97). SMT focuses more on the notion of "legibility conditions" whose core thesis is that the $C_{HL}$ (or otherwise, the narrow syntax) makes available the optimal way of relating an arbitrary set of lexical items to the interfaces, i.e. FL (i.e. the conceptual-intentional system) and PF (the conceptual-intentional system)" (Shormani, 2024: 4, see also Chomsky, 2005, 2008). An important generative notion related to SMT is *Search Space* (first appears in Chomsky's (2000)), which specifies feature valuation/interpretation mechanism between a probe *P* and a goal *G*. The former is usually functional and the latter a lexical constituent like nouns, adjectives, etc. *P's* features are usually unvalued/uninterpretable and *G's* usually valued/interpretable. Feature valuation/interpretation then takes place in such a way that the (active) *P* searches for a *G* in its *search space*. When both "meet", an *Agree* relation is established between both whose result is valuing/interpreting all *P's* unvalued/uninterpretable features including φ-features, Case, (in)definiteness (as far as Arabic DPs are concerned). With this in mind, given that syntactic ambiguity arises from structural configuration rather than lexical items themselves, disambiguation then consists in *Merge, Agree,* and feature valuation/interpretation among other operations/conditions.

From this perspective, syntactic ambiguity can be understood as a competition between alternative structural configurations or derivational possibilities: the same lexical material may participate in different hierarchical organizations, producing different interpretations (Chomsky, 2002). Chomsky (2021) reflects the continuing nature of this research program. Contemporary minimalist inquiry continues to address questions concerning *Merge, Agree, Locality, Phases, Feature Inheritance*, interpretation, and the computational architecture underlying syntactic objects (Chomsky, 2016). For Arabic, these questions remain empirically productive because the language provides extensive evidence concerning the interaction of syntax, morphology, Case, agreement, movement, and interpretation, including phenomena such as CSs, agreement, clause structure, and structural ambiguity.

## 2.2. Generative syntax and LLMs

The main idea, conception and working mechanism of LLMs perhaps originate in Chomsky's notions of human brain's (language faculty) "Generativity": from a finite number of words and rules, the human brain can "generate" an infinite number of sentences (Chomsky, 1957, et seq). the difference lies in arguing that LLMs' working mechanism is heavily built on statistical probabilities. However, LLMs acquire representations of linguistic patterns from large-scale textual data on which they are trained, thus paralleling the child's acquisition of L1 (from birth to 12 years) in which it receives tons of linguistic data (Shormani, 2014b). Perhaps one piece of evidence is the fact that LLMs' capacity lies within their working mechanism, which depends heavily on Neural Networking Algorithems (NNAs). The term "neural" implies "human", specifically human brain.

The relationship between AI and the human brain can be traced to the origins of artificial neural networks (ANNs), some of the earliest learning algorithms being conceived as computational abstractions of biological learning (Shormani, 2025), connecting this to "how learning happens or could happen in the brain" (Goodfellow et al., 2016: 14). ANNs' working mechanism to some extent tries to reproduce the brain's neural mechanisms and modern deep learning increasingly focuses on developing computational systems capable of learning complex representations and patterns from data. The emergence of TBMs represents a further development in this trajectory (Vaswani et al., 2017). This is also consolidated by NNAs, which further demonstrate the computing-biological correlation (Shormani, 2025). Today's TBMs could be thought as computational replicas of the human brain, their ability to construct context-sensitive representations has generated growing interest in their relationship to human language processing. In particular, research comparing neural language models with human neural and behavioral data has shown that representations learned by some language models can predict aspects of human brain activity during language processing, suggesting partial computational convergence between artificial language models and the human language system (Schrimpf et al., 2021; Caucheteux & King, 2022; Shormani, 2025). Such correlation, at least now, is not that clear, but coming days and future research could make such a kind of thought a reality, proving both entities' cognition (see also van Oostendorp and D'Alessandro, 2025).

The resulting research question is then not simply whether AraBERT "understands" Arabic, but more specifically: When generative syntax establishes two structurally legitimate analyses for an Arabic expression, can AraBERT distinguish between those competing analyses? This question provides the conceptual bridge between generative syntactic theory and LLMs, and establishes the theoretical basis for evaluating AraBERT as a computational system for Arabic syntactic ambiguity resolution. Another question this study strives to answer is: can generative syntax notions be integrated into LLMs' working mechanisms, and if so, to what extent can this integration improve LLMs' performance, specifically in handing a considerably complex structure, viz., ambiguity?

## 2.3. Deep Learning

Deep learning emerged as an extension of ANNs and is understood as "machine learning that achieves great power and flexibility by learning to represent the world as a nested hierarchy of concepts, with each concept defined in relation to simpler concepts, and more abstract representations computed in terms of less abstract ones" (Goodfellow et al., 2016: 8). Although AI research dates back to the 1940s, deep learning gained prominence particularly from the mid-2000s onwards (Goodfellow et al., 2016), with more recent developments further advancing neural network applications (Shormani, 2025). Different from earlier ML approaches such as Naïve Bayes (Metsis et al., 2006; Wickramasinghe & Kalutarage, 2021), Support Vector Machines (Hearst et al., 1998), Logistic Regression (Bagley et al., 2001), and Random Forests (Breiman, 2001), which relied substantially on handcrafted lexical, morphological, syntactic, and semantic features, DL learns representations directly from data. This reduces dependence on task-specific feature engineering, although deep neural networks introduce numerous hyperparameters, including hidden layers, learning rates, activation functions, batch sizes, and dropout rates, whose optimization commonly relies on validation-based and automated approaches (Wickramasinghe & Kalutarage, 2021). Early neural architectures for NLP, particularly recurrent neural networks and their gated variants such as LSTMs and GRUs, improved the modeling of sequential language data (Goodfellow et al., 2016), but their sequential processing was subsequently superseded by the Transformer architecture (Vaswani et al., 2017).

Transformers represented a major advancement through attention-based parallel processing, particularly self-attention and multi-head attention, which enable models to capture relationships among tokens across both local and long-range contexts (Vaswani et al., 2017). Building on this architecture, BERT introduced deeply bidirectional contextual representations through large-scale pre-training, establishing a powerful foundation for downstream NLP tasks (Devlin et al., 2019). Transformer-based models subsequently became the basis for a broad family of pretrained language models, including BERT, RoBERTa, XLNet, ELECTRA, and GPT, as well as language-specific variants (Annepaka & Pakray, 2025). These developments are particularly relevant to multi-label classification (MLC), in which an input may simultaneously receive multiple overlapping labels and therefore requires the model to learn multiple decision boundaries rather than a single mutually exclusive classification.

## 2.4. AraBERT

AraBERT is perhaps the most influential pretrained Transformer model developed specifically for Arabic NLP (Antoun et al., 2020). It was designed to address language-specific challenges associated with Arabic, including rich morphological variation, cliticization, and lexical sparsity. The model has been released in several versions (v0.1, v0.2, v1, and v2), with Base and Large configurations offering different levels of representational capacity for downstream NLP applications (Antoun et al., 2020, 2021, see also Alhussen & Shormani, 2026). The Base configuration comprises 12 Transformer encoder layers, a hidden size of 768, and 12 self-attention heads, whereas the Large configuration contains 24 encoder layers, a hidden size of 1024, and 16 self-attention heads. AraBERT uses a WordPiece-based tokenizer with an Arabic vocabulary designed to accommodate the morphological and cliticization characteristics of Arabic (Alhussen & Shormani, 2026. The model versions also differ in their preprocessing strategies: v0.1 and v0.2 were trained on raw, unsegmented Arabic text, whereas v1 and v2 use Farasa-based segmentation before tokenization, separating common prefixes, suffixes, and clitics to reduce sparsity and improve the representation of morphologically complex Arabic forms (Abdelali et al., 2016; Antoun et al., 2020; Alhussen & Shormani, 2026).

Following the BERT framework, AraBERT is pretrained primarily through masked language modeling, whereby masked tokens are predicted from their bidirectional contextual representations (Devlin et al., 2019; Antoun et al., 2020). Its pretraining draws on large-scale Arabic textual resources, including news articles, Arabic Wikipedia, OSCAR, and other web-crawled sources, amounting to tens of gigabytes of Arabic text (Antoun et al., 2020; see also Alkaabi et al., 2025; Alhussen & Shormani, 2026). These resources provide contextual representations that capture statistical regularities across diverse Arabic linguistic environments, making AraBERT suitable for downstream tasks involving morphosyntactic and contextual interpretation. In this study, this contextual modeling capacity is particularly relevant to structural ambiguity resolution: generative linguistic analysis defines the competing structural candidates, while AraBERT processes the candidate-conditioned input and learns to discriminate between the alternative interpretations on the basis of contextual and morphosyntactic evidence (Alhussen & Shormani, 2026). Thus, AraBERT provides Arabic-specific contextual representations derived from BERT-style Transformer pretraining. The ambiguity disambiguation task was formulated as binary sequence-pair classification.

### 2.5. Syntactic ambiguity

Syntactic (also called structural) ambiguity arises when a sequence of constituents permits more than one syntactic organization, attachment relation, or structural derivation (Shormani, 2014a). the study of syntactic ambiguity provides an important means of examining how linguistic form is mapped onto alternative structural representations and interpretations (McCarthy, 2002). A single overt expression permits two or more structurally distinct interpretations. The relevant distinction is, therefore, not simply semantic: the competing interpretations must correspond to alternative syntactic organizations. This perspective follows the generative view that surface strings may be associated with different hierarchical structures (Chomsky, 2004, 2007).

Structural ambiguity can be observed at two major levels: the nominal domain and the clausal domain. A familiar example from English is the phrase *French history teacher*, which allows at least two interpretations: (i) a French teacher who teaches history, and (ii) a teacher of French history (McCarthy, 2002: 71, see also Shormani, 2014a). A comparable type of ambiguity can also be found in Arabic, as illustrated in (2). This parallel suggests that ambiguity is not restricted to a particular language but may reflect a more general property of natural language (Chomsky, 2002). Arabic provides a particularly useful domain for examining such ambiguity because of its rich inflectional morphology, relatively flexible constituent ordering, agreement system, CS structures, and complex nominal configurations. Ambiguity may arise when the available morphosyntactic information is compatible with more than one structural analysis, particularly when a modifier or dependent can be associated with different constituents (see e.g., Hurford and Heasley, 1983; Othman et al., 2003; Piantadosi et al., 2012). In this way, resolving Arabic syntactic ambiguity requires sensitivity to hierarchical relations rather than reliance on surface word order or lexical association alone.

At the clausal level, ambiguity may arise from the possibility of assigning different syntactic relationships between a predicate and argument(s) associated with it. This can be illustrated by the sentence in (1) (from Shormani, 2014a: 9):

(1) ʔištarat-hu ʕabd-an
bought-she-him slave-acc
'She bought him a slave.'

According to Hurford and Heasley (1983 :128), ambiguity in sentences of this type results from the availability of two different sentence patterns. In such cases, the surface sequence of words does not by itself make it immediately clear which underlying structural pattern is intended. Thus, the sentence (1) can receive two distinct interpretations. The first interpretation corresponds to *ʔištarat ʕabd-an la-hu* 'she bought a slave for him', where *-hu* functions as an indirect object, while *ʕabd-an* functions as direct object of the verb given that the verb *ʔištarat* 'bought' is ditransitive. The second interpretation corresponds to *ʔištarat-hu ʕindama kān-a ʕabd-an* 'she bought him when he was a slave'. Both interpretations are structurally schematized in (2a & 2b), respectively.

(2) a. b.

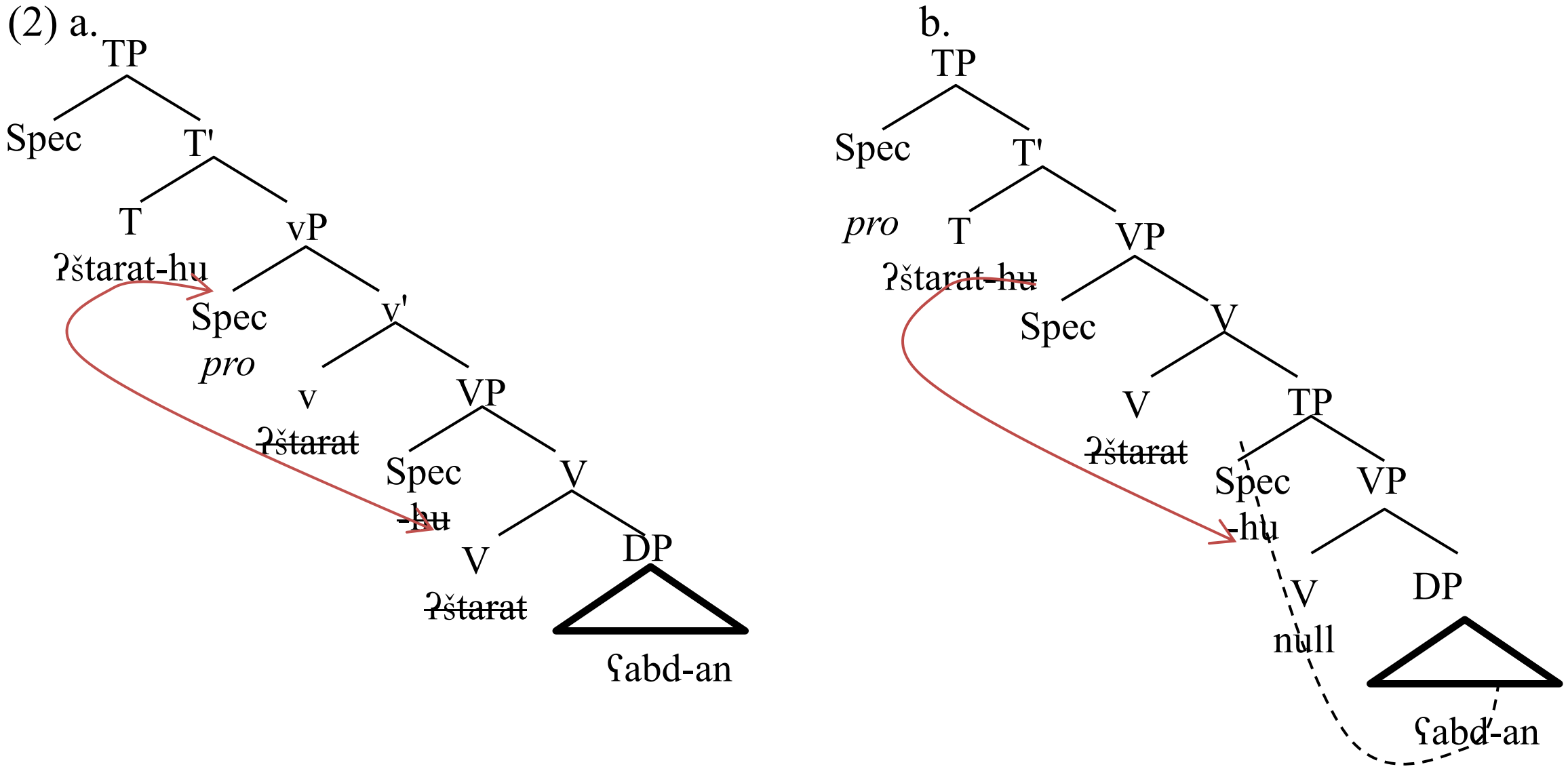

The two readings are associated with different structural configurations. The first can be represented as [...Obj1 + Obj2], involving two objects, whereas the second can be represented as [...Obj + (Obj Compl)], in which *-hu ʕabd-an* forms a small-clause-like constituent (indicated with a doted arow). Differently put, the second interpretation involves a predicative relationship between *-hu* and *ʕabd-an*, whereas this relationship is absent under the first interpretation. The ambiguity in (1), therefore, arises not from the lexical meanings of the individual words, but from the alternative syntactic relationships that can be established among them.

## 2.6. Syntactic ambiguity in Arabic DPs

In this section, we sketch structural ambiguity in Arabic. In Arabic, there are several types of DPs including CSs and nonCSs. simple and complex. The former is represented by (3a) and the latter by (3b), both of which do not involve ambiguity while (3c) does.

(3) a. al-kitāb-u l-jadīd-u
the-book -nom the-new-nom
'The new book'
b. kitāb-u ŧ-ŧālib-i
Book-nom the-student-gen
'The student's book'

c. al-kutubu wa d-dafātiru l-kabīratu
the-books and the-notebooks the-big
'the big books and notebooks'

(3c) is a nonCS coordinated DP whose ambiguity arises from the syntactic configuration in which al-kutubu 'books' and d-dafātiru 'notebooks' are related to the adjective al-kabīrat-u 'big'

Before addressing ambiguity in Arabic DPs, we need to address an important aspect of MSA. In this variety, unlike SA/ClA, the harakat ُ, َ and ِ do not show up on words. These harakat denote different cases, viz., nominative, accusative and genitive. Ambiguity in MSA appears perhaps more than in SA/ClA. Consider (4).

(4) kitāb-u ŧ-ŧālib-i l-kabīr-u SA/ClA
book-nom the-student-gen the-big-nom
'The student's big book'

b. kitāb-u ŧ-ŧālib-i l-kabīr-i SA/ClA
book-nom the-student-gen the-big-gen
'The book of the old student.

In both (4a) and (4b), there seems to be no ambiguity in SA/ClA. In (4a), for instance, it is the book which is big as both are nominative, but in (4b), it is the student which is big. However, there can be ambiguity, but iff the head N1 is genitive, say, a complement of a preposition as in (5)

(5) ʕitamad-tu ʕala [kitāb-i ŧ-ŧālib-i l-kabīr-i]
depended-I on [book-gen the-student-Gen the-big-gen]
'I depended on [the big book of the student/the old student's book]'

Structural ambiguity has received comparatively less attention in linguistic research compared to other phenomena (Shormani, 2014a). One reason lies perhaps in the fact that structural ambiguity does not originate in the lexical items themselves, but from the different ways in which those items can be syntactically organized within a phrase or sentence. Such alternative structural configurations are often less immediately apparent and therefore more difficult to identify (Elghamry, 2004; Al Daimi & Abdel-Amir, 199; Othman et al., 2003; Hurford & Heasley, 1983). In essence, structural ambiguity concerns the question of which constituent is syntactically associated with which other constituent within a given a piece of language.

Consider the following example, which illustrates structural ambiguity within the nominal domain.

(6) bayt ?axi l-kabīr
house brother.my the-big
'My older brother's house'

The DP bayt ?axi l-kabīr permits two possible interpretations as schematized in (7a & 7b).

(7) a

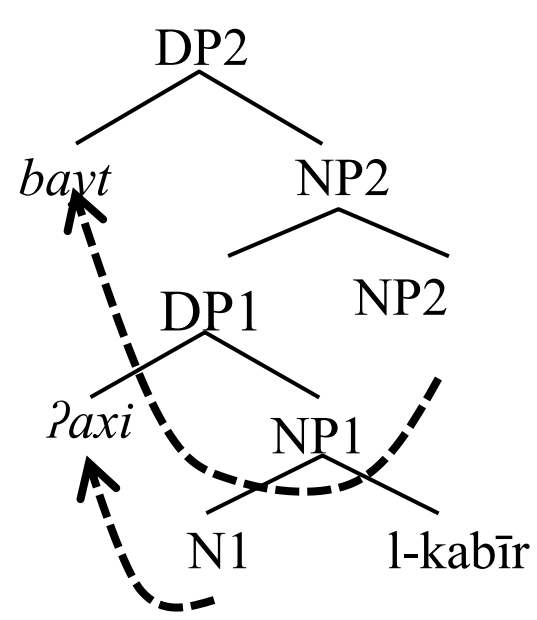

b.

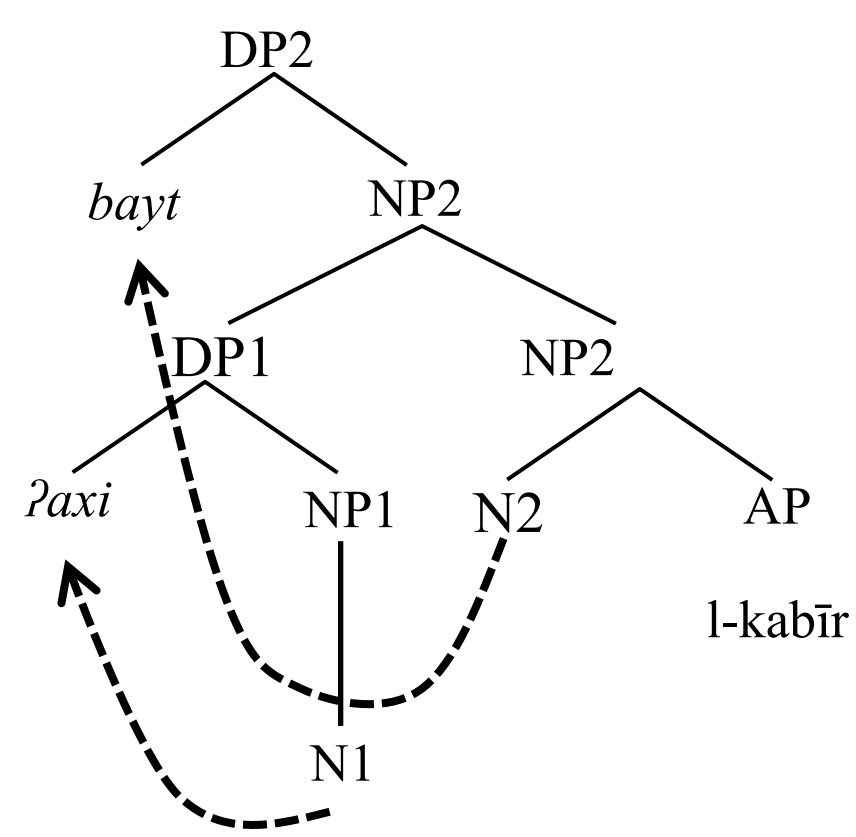

Under the first interpretation, the adjective *l-kabīr* modifies the head noun *bayt* 'house'. This meaning is shown in (7a). However, the adjective modifies the DPC *?axi* 'brother' in the second interpretation, as (7b) demonstrates. The ambiguity results from two alternative syntactic structures or hierarchical representations of the same sequence of constituents.

However, given that CSs (even nonCSs) are recursive (generative in nature), viz., the attachment can be multiple, be it to the head Ns or nonNs. Consider (8).

(8) a. qalam hibr ʔinjlizi kabīr
pen ink English big
'A big English ink pen'

This kind of ambiguity is schematically resolved in (9a & 9b)

(9) a. b.

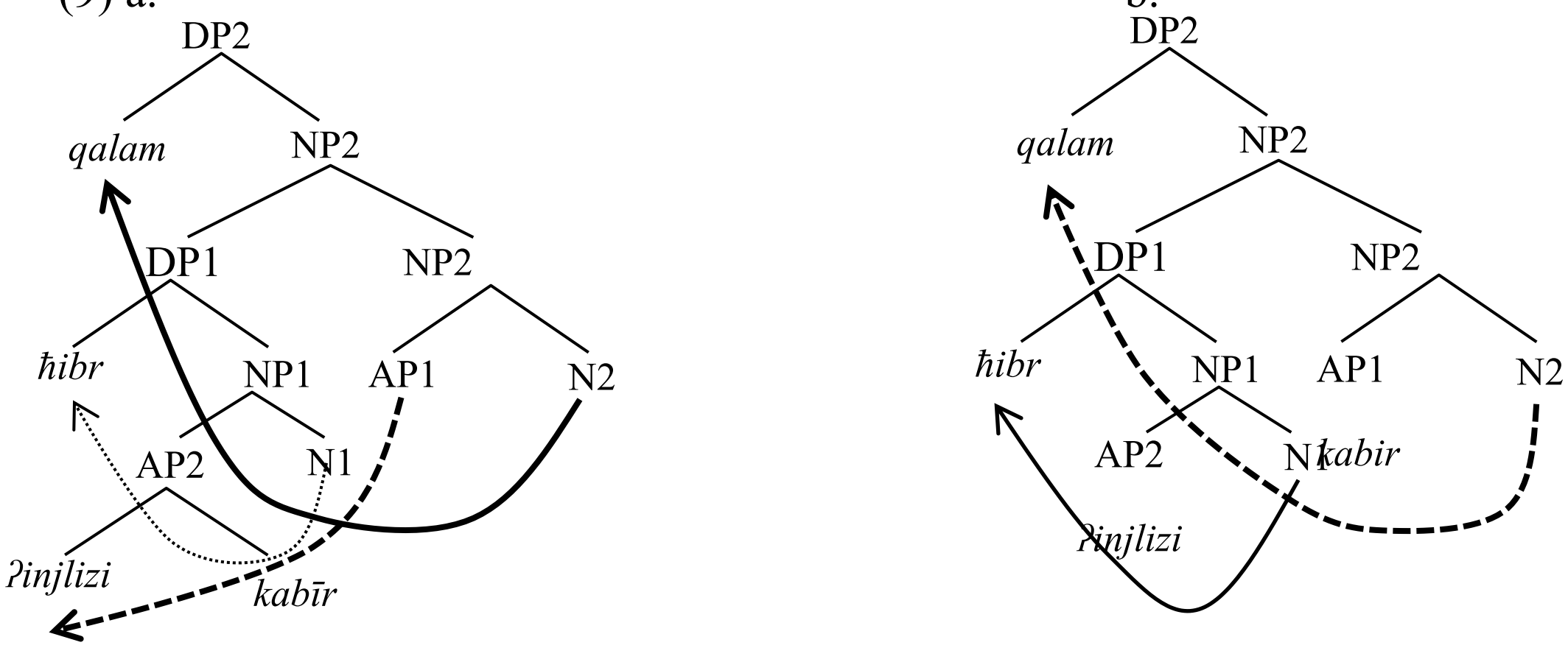

Thus, this study conceptualizes ambiguity as the interaction of three dimensions: (i) structural configuration, namely the hierarchical organization of constituents, (ii) morphosyntactic compatibility, including features such as (in)definiteness, φ-features, and Case, and (iii) interpretive consequence, whereby alternative structures yield different readings.

### 2.7. Related work

To the best of our know, no study has investigated this topic. However, there are several studies, which, in a way or another, are related to our study. We synthesize these in terms of the following themes: (1) syntactic and attachment ambiguity as a computational problem, (2) corpus-based and statistical approaches to attachment resolution, (3) neural and contextualized representation approaches, (4) Arabic syntactic processing with pretrained Transformer models, and (5) an emerging trend involving neuro-symbolic integration. These themes demonstrate the progression from manually designed linguistic and lexical features toward contextualized representations and provide the basis for investigating AraBERT as a model for Arabic attachment ambiguity resolution.

The first theme focusses on syntactic ambiguity and parsing, which has traditionally been regarded as a fundamental difficulty in natural language parsing because the correct syntactic structure cannot always be determined solely from the local sequence of words (van Gompel and Pickering 2007). Attachment ambiguity is a particularly important form of structural ambiguity. In PP attachment, for example, the same PP may plausibly modify either a preceding verb or a preceding noun. In this sense, a computational system must use information beyond the surface occurrence of the words to determine the intended interpretation. The importance of this problem is well established in computational linguistics. Belinkov (2014) describe PP attachment disambiguation as a longstanding challenge in syntactic parsing and emphasize the difficulty caused by lexical sparsity. Their work demonstrates that attachment decisions require representations capable of capturing syntactic and semantic properties of the words involved. Using Arabic and English datasets, they showed that learned word representations could substantially improve PP attachment prediction. Their neural model achieved 82.6% accuracy on the Arabic dataset, outperforming the Turbo and Charniak self-trained parsers, which obtained 76.7% and 80.8%, respectively.

The Arabic context makes syntactic ambiguity particularly significant. Abdelkader et al. (2024) conducted a corpus study of contemporary Arabic using a syntactically annotated subset of the Jordan comprehensive contemporary Arabic corpus containing more than 100,000 words. Their findings identify several sources of syntactic ambiguity, including the omission of diacritics, pro-drop constructions, flexible word order, constructions, agreement phenomena, and PP attachment. The authors specifically identify PP attachment as a challenging phenomenon for computational parsers attempting to determine the appropriate syntactic representation.

Another important theme is employing corpus-based, lexical, and statistical approaches to attachment resolution. For Arabic, Khan (2019) investigated PP attachment ambiguity using word-collocation information obtained from the arTenTen12 Arabic corpus through Sketch Engine. The study formulated the problem as deciding whether a PP attaches to the preceding noun (low attachment), the preceding verb (high attachment), or remains unclear. Human judgments were first used to construct labelled examples, after which collocational frequencies were used to derive rules for resolving attachment. The resulting approach achieved 83% precision and 88% recall on unseen examples. The limitations of purely lexical representations were also addressed in subsequent research using distributed word representations. Madhyastha (2017), for example, proposed a low-rank multilinear model for PP attachment based on products of word embeddings. This approach attempted to represent interactions among lexical features in a distributed vector space rather than relying exclusively on discrete lexical features. The results demonstrated that relatively simple models based on word embeddings could outperform more complex nonlinear architectures using the same lexical information and achieve competitive performance in out-of-domain settings.

A further theme concerns neural networks and contextualized representations for ambiguity resolution. Belinkov (2014) made use of introducing neural representation learning substantially, which changed the way syntactic ambiguity could be addressed. Belinkov provides an important example of this transition. The nonlinear architecture was initialized using word vectors learned from large amounts of unlabeled text and subsequently trained discriminatively for PP attachment. The authors also investigated dependency-based word representations and demonstrated that learned representations could capture information useful for syntactic attachment decisions. Their experiments on both English and Arabic showed that neural word representations could provide significant gains over traditional classifiers and existing parsers (cf. also Schnider et al., 2022).

Another theme that could be looked at here concerns Arabic syntactic processing and TBMs. For example, Al-Ghamdi et al. (2023) investigated the application of BERT-based pretrained models to Arabic dependency parsing. Their study compared nine Arabic BERT models, several finetuing strategies, and different encoding methods using three Arabic dependency treebanks: the Prague Arabic Dependency Treebank (PADT), the Columbia Arabic Treebank (CATiB), and the Classical Arabic Poetry Dependency Treebank (ArPoT). Their results showed that AraBERTv2 achieved the best performance across the three treebanks. The study also found that finetuning the higher layers of pretrained BERT models was important for capturing syntactic information, while adding additional neural layers could reduce performance.

However, there is an emerging theme focusing on neuro-symbolic trend and incorporating it into NLP models. This them is somehow related to our study, though different from ours in scope, purpose and content. As far we can tell, it is initiated by Chatzikyriakidis and Lappin (2026) who provided a systematic account of neuro-symbolic NLP, motivated by the complementary strengths and limitations of neural and symbolic approaches. They argue that neural models offer scalability and strong pattern-learning capabilities, while symbolic approaches can provide explicit structure, compositionality, reasoning, and greater transparency. Their principal contribution is a taxonomy that combines Kautz's classification of neuro-symbolic integration with Lappin's distinction between injective architectures, in which symbolic information is incorporated within neural processing, and federative architectures, in which neural and symbolic components remain separate but interact through an interface. Additionally, Mahouachi (2026) provided a more direct Arabic NLP precedent for the use of the term neuro-symbolic. The proposed MultiView+FuzzyRules framework combines contextual Transformer representations with static word embeddings and a differentiable fuzzy-logic feature layer. The symbolic component encodes domain-inspired linguistic indicators, including aggression, profanity, stylistic markers, and lexicon-based ratios, and uses fuzzy conjunction, disjunction, and negation to influence the neural model's predictions. The system is evaluated on Arabic offensive-language data spanning Tunisian, Egyptian, and Lebanese varieties, with additional experiments on Algerian dialectal abusive comments and Arabic YouTube comments. The authors evaluate both binary and multiclass offensive-language detection and report in-domain macro-F1 values of approximately 0.74–0.85 and cross-domain values of 0.63–0.76. Thus, the study demonstrates that symbolic linguistic or domain knowledge can be integrated with neural representations in an Arabic NLP system while addressing interpretability and dialectal variation.

### 2.8. Research gap

However, a specific gap remains at the intersection of neuro-symbolic NLP, generative syntax, and Arabic syntactic ambiguity resolution. As we have just seen, two studies have "nocked" this door by integrating neuro-symbolic for language classification. However, there is limited evidence on whether explicitly formulated generative-syntactic representations can be incorporated into a Transformer to resolve structurally defined ambiguity in Arabic. In particular, the use of hierarchical structure, *Merge, c-command, attachment relations*, and competing N1/N2 interpretations as candidate representations for neural ambiguity resolution remains insufficiently explored. This study addresses this gap by developing a generative-informed candidate-based framework in which generative linguistics defines the competing structural alternatives, human annotation establishes the appropriate interpretation, and AraBERT performs the contextual discrimination between them. Our contribution is sketched in what follows.

### 2.9. Our proposal

Given the complexity of ambiguity (see sections 2.5. & 2.6.), it seems that TBMs would certainly encounter difficulty in processing and/or learning this phenomenon. Thus, in complex NPs [NP $N_1$ [NP $N_2$]] + Adj, like (2-9) an adjective can attach high to the governing head noun ($N_1$) or low to the genitive complement ($N_2$). Similarly, PP attachment within complex determiner phrases like [DP $DP_1$ [DP $DP_2$]] + PP and coordinate noun modification like [$N_1$ and $N_2$] + Modifier present inherent structural ambiguities. In such complex DPs involving coordinate noun modification ($DP_1$ and $DP_2$ + Modifier), the algorithm must decide whether the modifier scopes over the entire coordinated constituent (High Attachment) or restricts only the second noun (Low Attachment). Given that standard transformer tokenizers process words without explicit phrase-boundary markers, self-attention cannot dynamically "group" $DP_1$ and $DP_2$ into a single structural node without explicit sequence-pair formatting or syntactic constraints.

More importantly, Symmetrical Attention Matrices (SAM) vs Asymmetrical C-Command (ACC) (cf. Kayne, 1994) provide an enough room for generative syntax to dictate that constituent relationships are asymmetrical- a governing head c-commands its complement, but the reverse is not true. Softmax-normalized self-attention layers calculate pairwise token interactions ($QK^T$) that struggle to enforce directional, bounded hierarchical dominance. Without structural biases, attention heads distribute probability mass across all contextually related tokens, leading to "over-smoothing" where the model fails to isolate whether an adjunct modifies the outer DP node or an inner embedded DP node. DL algorithms like standard self-attention models compute token interactions primarily over linear sequence positions, which often leads to structural failure when resolving these phrase-level ambiguities (cf. Abdeahad et al., 2026). In MSA DPs, surface lexical co-occurrence alone is frequently insufficient to determine attachment heights (High vs Low attachment). Generative syntax explicitly defines these phrase relationships as hierarchical trees generated by formal operations like *Merge* (see (1) through (9) above). Providing structured sequence inputs such as candidate attachment pairs explicitly formatted alongside the main clause and feeding TBMs like AraBERT, forces their cross-attention mechanism to evaluate structural argument satisfaction directly, prioritizing hierarchical constituent depth over mere linear token proximity.

Given our "search space" story (see section 2.1.) in generative minimalism, a similar notion is also found in LLMs, almost perhaps with the same working mechanism. TBMs/LLMs' *Search Space* represents the combinatorial universe of all possible token sequences that the network can generate at any given step (see e.g., Lin et al., 2026). Unlike symbolic parsers that traverse a bounded, rule-governed search space defined by formal grammar constraints (such as valid syntactic trees formed by *Merge* operations), an LLM operates over an unbounded vocabulary-level search space. At each decoding step, the model computes a continuous probability distribution over its entire vocabulary, mapping contextual attention vectors to a high-dimensional space of potential next tokens. However, given the recursive nature of LLMs' (also human bran's) "generativity", and due to the massive data LLMs have been trained on, LLMs may encounter vast "search space", navigating which relies on decoding strategies like *greedy search, beam search, or nucleus sampling* (Lin et al., 2026). In neuro-symbolic and candidate-conditioned architectures, this search space is explicitly constrained by injecting pre-formulated structural alternatives into the input, effectively channeling the LLM's attention mechanism to evaluate candidate paths within a bounded structural space rather than generating unrestrained token sequences.

Thus, feeding TBMs with explicit syntactic assumptions effectively constrains the model's search space during finetuning. Unconstrained finetuning forces the TBMs to jointly learn Arabic morphology, lexical semantics, and phrase structure simultaneously from limited gold-standard annotations. By feeding AraBERT syntactically structured input pairs, we could "inject" a formal structural prior. The model no longer needs to deduce where the syntactic juncture lies; instead, its attention heads are directed to evaluate the syntactic viability of specific candidate constituent configurations. This syntax-guided approach directly mitigates the challenge of morphosyntactic feature sparsity in Arabic NLP. MSA features a rich agreement system where φ-feature, Case, and definiteness govern syntactic validity across distant clause/phrase elements. Generative approaches, specifically minimalism model these features as formal uninterpretable and interpretable feature sets that must be checked through structural configurations. Delimiting candidate constituent heads within the input sequence guides multi-head self-attention towards agreement-bearing heads and their dependents, resulting in robust generalization across unobserved vocabulary and complex phrase structures.

To repeat a point stated earlier, given the structural complexity of syntactic ambiguity, TBMs including AraBERT would certainly encounter difficulty when interpretation depends on hierarchical relations that are not explicitly represented in the input. In complex Arabic DPs such as [NP $N_s$ [NP $N_s$]] + Adj(s), an adjective may attach to the higher noun ($N_1$) or the embedded (genitive) noun ($N_2$). Similar ambiguities arise in [DP $DP_1$ [DP $DP_2$]] + PP configurations and in [$N_1$ and $N_2$] + Modifier(s), where a modifier may scope over the entire coordination or only over the second conjunct (cf. Mohammad 2000). Such cases cannot always be resolved from lexical co-occurrence or linear proximity alone because the competing interpretations differ in hierarchical constituency and attachment. Generative syntax explicitly represents these alternatives through hierarchical structure, *IM*, *C*-command, *Agree,* and attachment relations, through which ambiguity is easily resolved. We propose that this ambiguity is resolved through a minimalist notion known as Feature Sharing Mechanism (FSM) (see Danon 2013; Shormani 2016, 2017a & b). FSM works in such a way that, due to agreement in Case, definiteness, and φ-features, the head N/D constitutes a permeant link with all the internal head Ds/Ns, and when the head D/N's unvalued/uninterpretable features get valued/interpreted, it (the head D) values all the unvalued/uninterpretable features in the link via *Agree*. Simply put, in Arabic multiple DPs/CSs constitute morphosyntactic agreement across nested modifiers, and if the head D/N (and all nested Ns/Ds), agrees with a modifier, then ambiguity gets disambiguated simply because all the nested Ds/Ns will agree with the same modifier, be it adjectival, nominal, or clausal.

Given this FSM, whatever nested (embedded) DPs/CSs, i.e., intervening genitive complements, when the head D's features are interpreted by a valuing noun, ambiguity gets resolved. If this is on the right track, and if we can "feed" AraBERT with these agreement features, then its self-attention heads can learn these features, hence disambiguate the meaning by linking N1 to its modifier(s), or N2 to its modifier(s). The proposed framework integrates these theoretically motivated distinctions with neural contextual modeling, hence not requiring the Transformer to infer the relevant structural alternatives from raw text alone. For each ambiguous construction, generative analysis defines competing candidate structures, while AraBERT receives these candidates as structured input and determines which interpretation is supported by the context. The framework is thus generatively informed and neuro-symbolic: symbolic structural knowledge defines the alternatives, human annotation establishes the appropriate interpretation, and the Transformer performs neural discrimination between them. Our proposal is schematized in Fig 1.

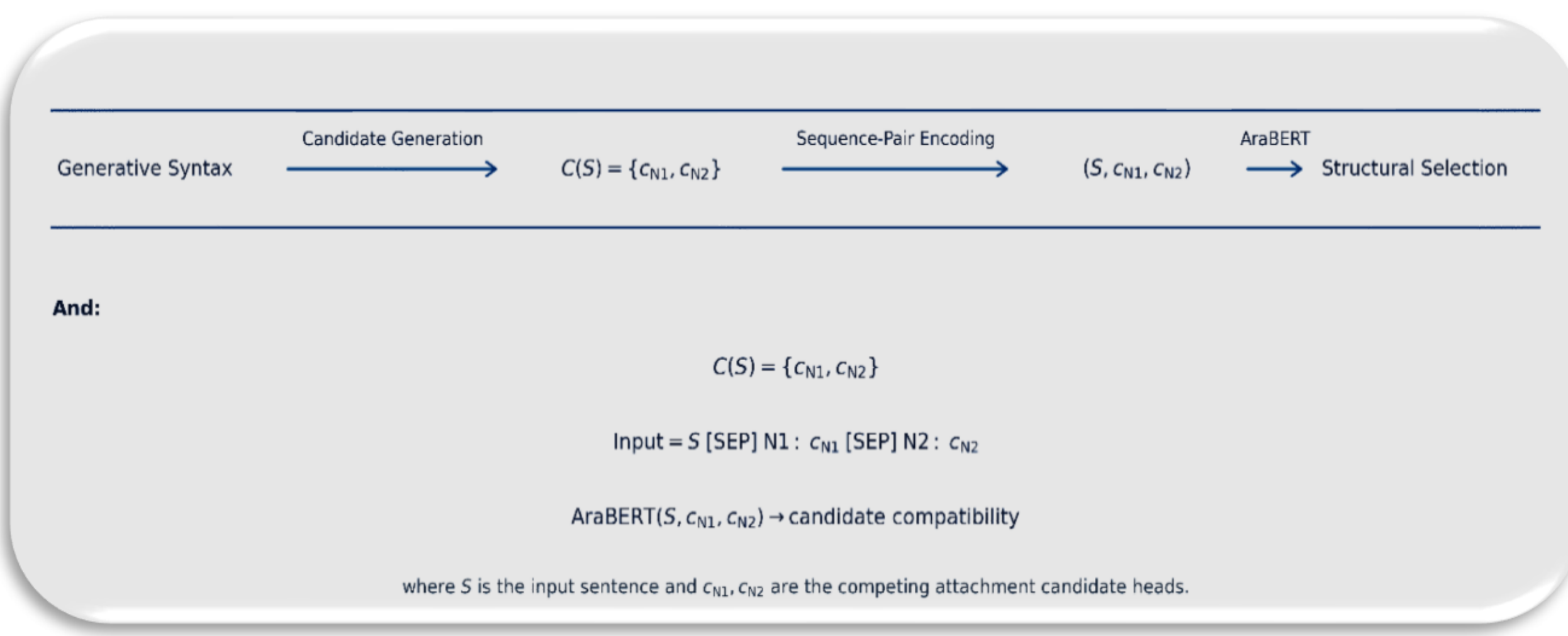

***Fig 1: Candidate-conditioned phrase-pair encoding pipeline***

Fig 1 summarizes the candidate-conditioned phrase-pair encoding framework designed to resolve structural syntactic ambiguity using AraBERT. In the top pipeline, the input text (s) represents the overall input sentence/utterance containing an ambiguous phrase *s*, which is processed by a generative syntax module during Candidate Generation to construct a set of valid candidate structural analyses, $C(S) = (c_{N1}, c_{N2})$, representing distinct syntactic interpretations such as high and low attachment. Next, through Phrase-Pair Encoding, the input sentence *S* and the competing candidate attachment heads ($c_{N1}$ and $c_{N2}$) are formatted into a single structured input sequence

$$\text{Input} = S\text{[SEP] N1: } c_{N1}\text{[SEP] N2:}c_{N2} \qquad (1)$$

This joint sequence representation ($S$, $c_{N1}, c_{N2}$) is then fed into AraBERT for Structural Selection, where the model evaluates the contextual and semantic compatibility across candidate attachments (AraBERT($S$, $c_{N1}$, $c_{N2}$) ➔ candidate compatibility).

By scoring each structured input directly, the model selects the candidate analysis that exhibits the highest structural compatibility within the given Arabic sentence context. To avoid treating ambiguity resolution as an unconstrained syntactic parsing task, this study redefines the problem as candidate-conditioned structural selection. Thus, in our proposal the neural model is not forced to predict or generate an unrestricted parse tree from scratch, but rather a generative syntactic component pre-defines the explicit search space $C(S) = (c_{N1}, c_{N2})$ ➔ $C(\text{S}) = \{ c_{N1}, c_{N2}\}$ for each ambiguous phrase. This setup constructs two structurally valid competing interpretations, representing high-attachment ($c_{N1}$) and low-attachment ($c_{N2}$) configurations, and allowing AraBERT to function as a discriminative scorer trained directly against human gold syntactic judgments (as we will see shortly).

## 3. Methodology

### 3.1. Study design

Given the scope and purpose of this study, we adopt a mixed-method approach framed within linguistically motivated candidate-construction stage with a data-driven contextual classification stage. This design reflects a progression from traditional symbolic approaches to neural representation learning. Probabilistic syntactic models such as Probabilistic Context-Free Grammars (PCFGs) represent alternative derivations explicitly and assign probabilities to competing parse trees (Scicluna & de La Higuera, 2016). Similarly, graph-based dependency parsers represent alternative dependency structures and select the highest-scoring valid tree (Nivre & McDonald, 2008). These approaches provide explicit structural representations and principled mechanisms for ranking alternatives. Transformer models instead learn contextual representations directly from large-scale data. The Transformer architecture uses self-attention to model relationships among tokens across a sequence (Vaswani et al., 2017), while BERT learns bidirectional contextual representations through masked language modeling (Devlin et al., 2019). Although masked language modeling does not explicitly construct syntactic trees, contextual representations can encode information relevant to hierarchical and morphosyntactic structure (see also Cowper et al., 2019).

### 2.2. Operationalizing syntactic ambiguity

Syntactic ambiguity was operationalized as a configuration in which a single overt expression permits two structurally distinct interpretations that are both compatible with the available linguistic evidence. Identification, therefore, did not depend simply on the presence of a particular construction. Instead, candidate instances had to satisfy three criteria: i) the existence of structurally distinct alternatives, ii) morphosyntactic compatibility with those alternatives, and iii) a meaningful interpretive contrast between them. The procedure was particularly suited to Arabic because of its rich inflectional system, relatively flexible constituent ordering, CS structures, agreement patterns, and complex nominal configurations. Morphological information was used to verify candidate structures and exclude analyses that were inconsistent with agreement, case, or other grammatical constraints. However, morphology was treated as evidence for structural analysis, but not, say, a substitute for syntax.

The ambiguity inventory included three principal configurations: CSs, NonCSs, and Coordinated DPs. CSs' ambiguity involved competing interpretations within complex structures, where a modifier or dependent may be associated with different nominal constituents, viz., head N1, or (genitive) head N2. NonCSs or Coordinated DPs' ambiguity involved alternative structural groupings or scopes of coordinated constituents. Competing attachment ambiguity involved a constituent that could plausibly be attached to a higher verbal or clausal domain or to a lower nominal/embedded domain. This formulation permits the same binary decision scheme to be applied across different ambiguity types while preserving their syntactic differences (cf. e.g. Yang et al., 2011).

### 3.3. Data preparation

The experimental corpus was constructed from two complementary resources: the UD Arabic-PADT treebank and WebFine-2. UD Arabic-PADT provides syntactically annotated MSA sentences in Universal Dependencies format, including head-dependent relations and morphosyntactic information. WebFine-2 supplements this material with naturally occurring web Arabic, increasing lexical and contextual diversity. The two resources were normalized to a common representation before candidate extraction. The resulting corpus contained 44637 sentence-level instances after normalization, deduplication, and structural filtering. The computational unit was the complete sentence, because syntactic interpretation depends on sentence-wide contextual information; however, the linguistic target of ambiguity was a phrase or DP syntactic constituent, not the sentence as a whole. The ambiguity inventory included CSs, NonCS, Coordinated DPs, and competing attachment relations.

For each qualifying instance, two candidate interpretations were constructed and represented explicitly as alternative structural analyses. Candidate N1 generally represented the higher structural interpretation, whereas Candidate N2 represented the competing local or embedded interpretation. The candidate representation was standardized so that the model encountered comparable alternatives across sentences of different lengths and syntactic complexity.

### 3.4. Annotation scheme

To establish a gold-standard benchmark, a stratified random sample of 1000 sentences was drawn from the processed corpus for human annotation. The sample was designed to mirror the principal ambiguity categories and structural configurations identified during corpus construction. This sample served as the primary annotation set for establishing task-specific ground-truth judgments. All authors constituted the expert annotation panel, collectively reviewing and establishing consensus on the annotation guidelines.

The human annotators assigned a binary target label corresponding to the two explicitly constructed candidate interpretations:

- Label 0 — High / VP Attachment ($c_{N1}$): The ambiguous constituent attaches to the matrix verb or higher structural domain, corresponding to Candidate N1.
- Label 1 — Low / NP / Embedded Attachment ($c_{N2}$): The ambiguous constituent attaches to a local noun phrase or embedded structural domain, corresponding to Candidate N2.

For CSs and coordinated constructions where the structural contrast does not strictly reflect an attachment relation in the traditional sense, the same binary paradigm was maintained: annotators selected the functionally appropriate interpretation from the two candidate options presented. The resulting judgments established a task-specific human reference standard. This reference standard is distinct from treebank-derived labels because the gold annotation directly evaluates the explicit binary candidate competition rather than simply replicating the syntactic representations stored in the source treebanks.

### 3.5. Candidate-based syntactic representation

The central methodological contribution rests on formulating ambiguity resolution as a competition between two explicit candidate structures rather than requiring AraBERT to generate or recover an unconstrained parse tree from scratch. For each sentence *S*, the model receives the sentence alongside its two candidate interpretations, viz., $(S, c_{N1}, c_{N2})$.

The disambiguation task is expressed as:

$$G(S, c_{N1}, c_{N2}) \in \{0, 1\} \quad (2)$$

where G denotes the human gold judgment, and prediction is given by:

$$\hat{G}(S, c_{N1}, c_{N2}) = f\mathrm{AraBERT}(S, c_{N1}, c_{N2}) \quad (3)$$

The training objective is to maximize agreement between model predictions ($\hat{G}$) and $G$. This formulation establishes a strict separation between two primary research components:

1. The linguistic component defines the candidate set $C(S) = \{S, c_{N1}, c_{N2}\}$ containing the structurally and morphosyntactically plausible interpretations derived from generative syntax principles.

2. The computational component Evaluates whether AraBERT can leverage sentence-level context $S$ to score candidate compatibility between $c_{N1}$ and $c_{N2}$.

The model success cannot be attributed simply to learning a shallow lexical association; the relevant options are explicitly defined as competing syntactic configurations.

### 3.6. Tokenization and normalization

Following candidate extraction, data preparation involved multi-stage morphological and structural verification. Arabic verbal morphology was parsed to verify agreement features, including person, number, gender, and inflectional affixes. Functional and verbal markers like *qad*, *sawfa*, *lan*, and *lam*, as well as *wāw al-jamʕ* were rigorously validated. In verbless nominal clauses, *yakūn* 'be' was designated as the default high structural head whenever required by the candidate representation.

Spurious candidate pairs were systematically removed if they originated from invalid particle segmentation, duplicated functional tokens, tokenization or morphological segmentation artifacts, structurally uninformative constituents, or material irrelevant to the core syntactic contrast. For example, adverbial modifiers such as *baʕīdan* were excluded when they did not interact with the competing structural interpretations. Identical normalization and filtering criteria were applied uniformly across all data partitions to maintain strict consistency.

### 3.7 Finetuning and optimization

Model fine-tuning utilized the pretrained aubmindlab/bert-base-arabertv02 checkpoint (Antoun et al., 2020). The complete sentence *S* served as the primary contextual sequence, with the two candidate structural representations incorporated directly into the model input. Specifically, each input was formatted as the preprocessed sentence followed by the two candidate representations, using the structure:

$$\text{Input} = S\text{[SEP] N1:}c_{N1}\text{[SEP] N2:}c_{N2} \qquad (1')$$

Arabic text was preprocessed using the AraBERT preprocessing pipeline, including Farasa-based morphological segmentation, before tokenization with the corresponding AraBERT tokenizer. The maximum input sequence length was set to 128 tokens. The same preprocessing, candidate representation, and tokenization procedures were applied consistently across the training, validation, and evaluation partitions.

The model extracted the contextual representation corresponding to the classification token ($h_{[CLS]}$), and estimated the probability distribution across the candidate classes as follows:

$$P(y \,/\, (S, c_{N1}, c_{N2}) = Softmax(W \cdot h_{[CLS]}) \qquad (4)$$

where $W$ represents the weight matrix of the binary classification head. Fine-tuning adjusted AraBERT's parameters so that its contextual representations increasingly aligned with the human-established syntactic interpretations.

AraBERT was finetuned using PyTorch and the HuggingFace Transformers framework. Model training optimized a binary cross-entropy loss within an equivalent two-class softmax classification setting using the AdamW optimizer. Optimization hyperparameters included a learning rate set to$2 \times 10^{-5}$ (or 0.00002), weight decay of 0.01, gradient-norm clipping capped at a maximum threshold of 1.0, and mixed-precision FP16 execution. A linear learning-rate decay schedule with a warm-up phase was employed to stabilize optimization across training epochs. Model checkpoints were evaluated at the end of each epoch on the validation partition, and the optimal checkpoint was selected based on the highest validation weighted F1 score.

The annotated dataset was partitioned using a stratified training, validation, and evaluation design. The training set was used for parameter updates, the validation set for model selection and early stopping monitoring, and the held-out evaluation set strictly for final performance assessment. Evaluation data were kept entirely isolated from parameter tuning and checkpoint selection. Recall that the initial extraction pool of 44637 raw sentence-level instances represented general Arabic text prior to ambiguity filtering. Through multi-stage structural filtering, morphological noise removal (Section 3.6), and expert annotation verification (Section 3.4), non-ambiguous and noisy instances were systematically removed. This pipeline yielded a final verified human-ground-truth corpus of 6696 target instances. The target class distribution contained 4881 High/VP attachment cases ($c_{N1}$; 72.89%) and 1815 Low/NP/Embedded attachment cases ($c_{N2}$; 27.11%). The computational pipeline was implemented in Colab (Python) using PyTorch, HuggingFace Transformers, scikit-learn, pandas, and Farasa. Rigorous reproducibility was ensured through a fixed random seed (SEED = 42), explicit model identification (aubmindlab/bert-base-arabertv2), standardized preprocessing and candidate-construction procedures, fixed hyperparameters, and consistent data partitioning. The methodological design thereby fully specifies both the linguistic principles used to construct the competing syntactic alternatives and the computational parameters used to train and evaluate AraBERT.

### 3.10. Methods of Analysis

Model performance was evaluated using accuracy, macro-precision, macro-recall, macro-F1, and weighted F1, together with class-specific precision, recall, and F1 for the two syntactic interpretations. Let *TP*, *TN*, *FP*, and *FN* denote true positives, true negatives, false positives, and false negatives, respectively. Then, overall accuracy was calculated as:

$$Accuracy = \frac{TP + TN}{TP + TN + FP + FN} \quad (5)$$

For the two-class task, macro-precision and macro-recall were calculated by averaging the class-specific values across the label set Y:

$$Precision = \frac{1}{|Y|} \sum_{c \in Y} \frac{TP_c}{TP_c + FP_c} \quad (6)$$

$$Recall = \frac{1}{|Y|} \sum_{c \in Y} \frac{TP_c}{TP_c + FN_c} \quad (7)$$

The macro-F1 score was then calculated as the mean of the class-specific F1 scores:

$$F1 - Score = \frac{1}{|Y|} \sum_{c \in Y} \frac{2 \cdot Precision_c \cdot Recall_c}{Precision_c + Recall_c} \quad (8)$$

Because the two classes were not equally represented, weighted F1 was additionally reported, with each class weighted according to its proportion in the evaluation set:

$$Weighted\,F1 = \sum_{c \in Y} \frac{n_c}{N} F1_c \quad (9)$$

where $n_c$ is the number of instances belonging to class cc and NN is the total number of evaluation instances.

In addition to these aggregate measures, a confusion matrix was used to examine the directionality of classification errors between N1 (High/VP Attachment) and N2 (Low/NP/Embedded

Attachment). Model logits were converted into class probabilities using the softmax function, and the predicted class and associated confidence were retained for subsequent analysis. This made it possible to distinguish low-confidence uncertainty from high-confidence misclassification.

The evaluation comprised 6696 human-ground-truth instances, including 4881 N1 cases and 1815 N2 cases. The confusion matrix, therefore, provided a direct assessment of how effectively the model distinguished the two competing syntactic interpretations and whether errors were systematically biased toward one structural analysis. Analysis was conducted at three levels: i) aggregate metrics measured overall classification performance, ii) class-specific and ambiguity-type analyses examined whether performance differed between High/VP and Low/NP/Embedded interpretations and across CS, coordination, and competing-attachment configurations, and iii) qualitative error analysis examined structurally difficult cases, lexical and morphological properties, contextual demands, and possible preprocessing or candidate-construction artifacts. Particular attention was given to high-confidence errors, as these may reveal systematic tendencies in the model's interpretation of competing syntactic structures rather than simple uncertainty.

## 4. Results

This section presents the study results. We first tabulate these results and then analyze them.

**Table 1: Arabert training and validation**

| Epoch | Training Loss | Validation Loss | Accuracy | Precision | Recall | F1 Binary | F1 Weighted |
|---|---|---|---|---|---|---|---|
| 1 | 0.179989 | 0.134209 | 0.961768 | 0.957421 | 0.894318 | 0.924794 | 0.961339 |
| 2 | 0.118960 | 0.118430 | 0.971326 | 0.988778 | 0.901136 | 0.942925 | 0.970884 |
| 3 | 0.088660 | 0.109212 | 0.976404 | 0.992015 | 0.917614 | 0.953365 | 0.976100 |
| 4 | 0.087190 | 0.106982 | 0.977599 | 0.995690 | 0.918750 | 0.955674 | 0.977301 |

Table 1 presents the training and validation performance of AraBERT across four epochs. Training loss decreased from 0.1800 in Epoch 1 to 0.0872 in Epoch 4, while validation loss declined from 0.1342 to 0.1070. At the same time, accuracy increased from 96.18% to 97.76%, and precision increased from 95.74% to 99.57%. Recall improved from 89.43% to 91.88%, resulting in an increase in binary F1 from 92.48% to 95.57%. Weighted F1 followed the same pattern, increasing from 96.13% in Epoch 1 to 97.73% in Epoch 4. Thus, Epoch 4 produced the strongest training and validation performance across all reported metrics.

**Table 2: Arabert performance on the unseen set**

| Training Loss | Validation Loss | Epoch | Accuracy | Precision | Recall | F1 Binary | F1 Weighted |
|---|---|---|---|---|---|---|---|
| 0.087190 | 0.142557 | 4 | 0.968787 | 0.991432 | 0.892562 | 0.939403 | 0.968252 |

Table 2 displays AraBERT's final evaluation performance achieving an accuracy of 96.88%, precision of 99.14%, recall of 89.26%, binary F1 of 93.94%, and weighted F1 of 96.83%. Class-level results show substantially higher recall for N1 (99.71%) than for N2 (89.26%), while precision was high for both classes (96.15% and 99.14%, respectively).

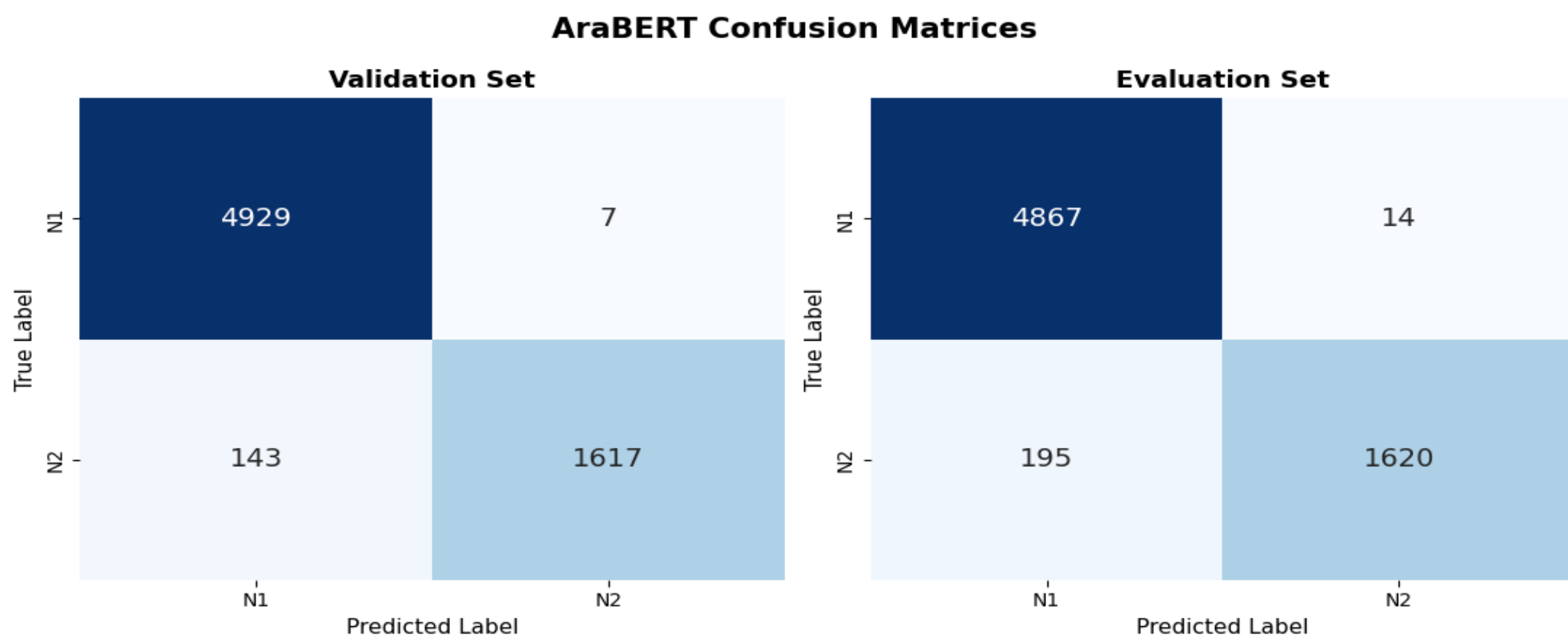

*Fig 2: Confusion matrices*

Fig 2 illustrates the classification distribution of AraBERT across the Validation Set (left) and the Evaluation/Unseen Set (right), High Precision for N2. The model exhibits exceptional precision on class N2, making only 7 false positive predictions on the validation set and 14 on the evaluation set. The primary source of error stems from class N2 instances being misclassified as N1 (143 cases in validation and 195 cases in evaluation). This reflects the underlying class distribution in the dataset while maintaining strong overall accuracy above 96% across both sets.

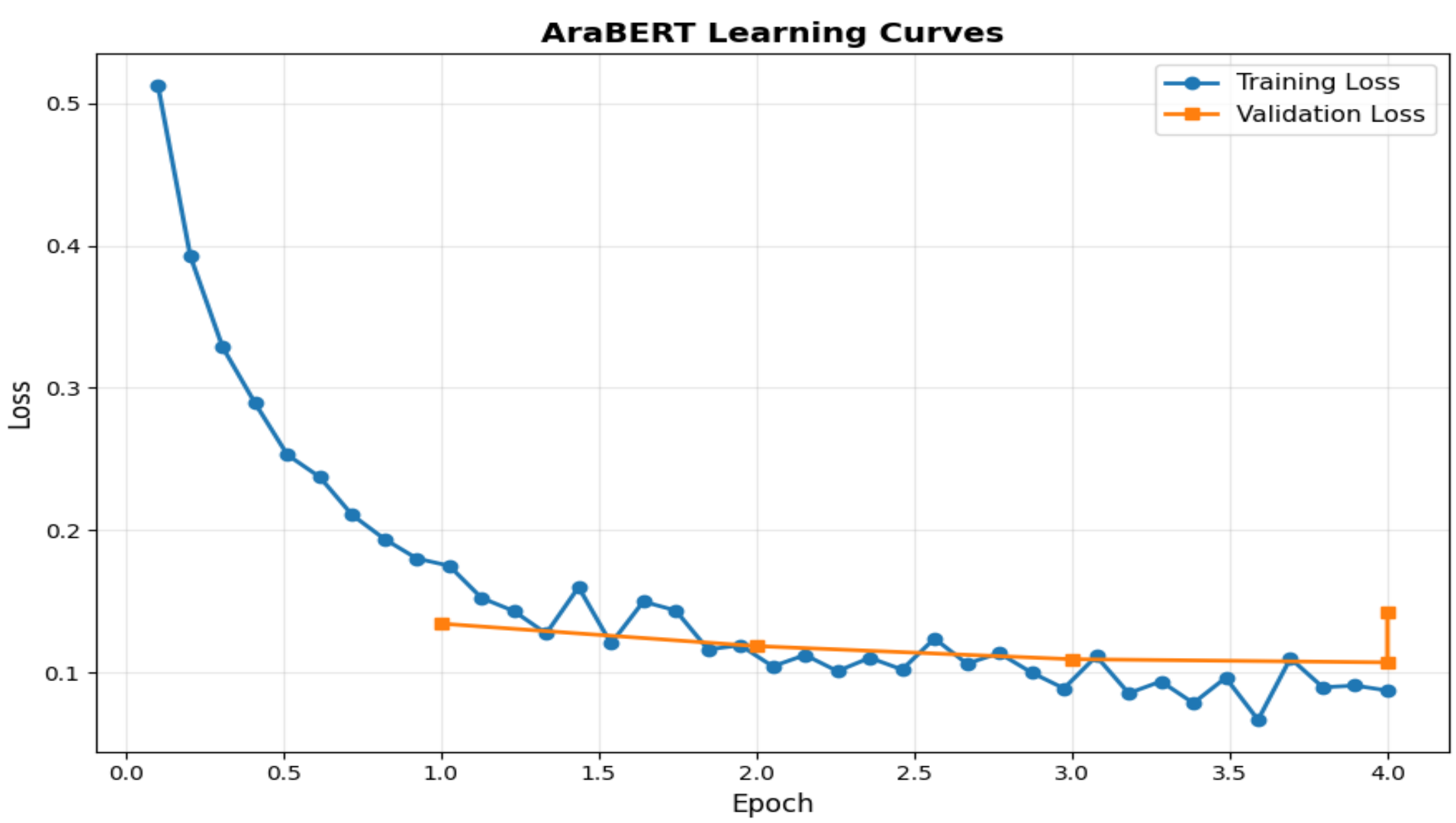

*Fig 3: Learning curves*

Fig 3 displays the optimization trajectory over the 4 finetuning epochs. Training Loss (Blue) shows a steep, continuous drop per training step, starting from over 0.50 and steadily stabilizing around 0.087, demonstrating effective convergence. Validation Loss (Orange): Decreases smoothly across the four epochs from 0.1342 down to 0.1070. The final data point reflects the evaluation performance on the unseen set (0.1426), confirming strong generalization without significant overfitting.

## 5. Discussion

The empirical findings strongly validate our central thesis: while standard Transformer architectures operate primarily over linear sequence positions, injecting formal generative priors through structured candidate pairs ($S$, $c_{N1}$, $c_{N2}$) successfully constrains the model's *search space*. Across four training epochs, AraBERT showed consistent improvement in both training and validation performance, with training loss declining from 0.1800 to 0.0872, validation loss decreasing from 0.1342 to 0.1070, and validation accuracy increasing from 96.18% to 97.76%. More importantly, performance remained high when the final model was evaluated on the unseen set containing 6696 instances with human-established ground truth. The model achieved 96.88% accuracy, 93.94% binary F1, and 96.83% weighted F1, with a macro-F1 of 95.92%. The minimal gap between validation performance at Epoch 4 and held-out evaluation performance highlights the model's strong generalization capacity when presented with unobserved vocabulary in deeply nested structures.

Standard multi-head self-attention mechanisms compute pairwise token interactions via $QK^T$, whose inherent limitations make them prone to "over-smoothing" and structural failure when resolving phrase-level ambiguities. By formatting the input as explicit candidate interpretations rather than forcing unconstrained generation, AraBERT was provided with a structural boundary prior. The model no longer had to independently induce abstract phrase-structure trees or recover latent constituent nodes from surface text alone. Instead, attention mechanisms were guided to evaluate structural argument satisfaction directly. By offloading constituent boundary discovery to the input formatting, AraBERT effectively allocated its parameters toward checking long-range agreement features (gender, number, case, and definiteness) across distant clause elements rather than struggling to group sequence tokens into hierarchical phrases.

At the same time, the class-specific results reveal that the two structural alternatives were not equally easy for AraBERT to identify. The model correctly classified 4867 of the 4881 N1 cases, yielding a recall of 99.71% for the High/VP or higher-attachment interpretation. By contrast, 1,620 of the 1815 N2 cases were correctly identified, corresponding to a recall of 89.26% for the Low/NP/embedded interpretation. The confusion matrix shows that 195 N2 instances were misclassified as N1, whereas only 14 N1 instances were classified as N2. Precision results reinforce this interpretation: N2 achieved a precision of 99.14%, meaning that when AraBERT predicted the low/embedded interpretation, that prediction was rarely incorrect. N1 showed very high recall (99.71%) but lower precision (96.15%), reflecting the 195 N2 instances that were absorbed into the N1 category.

This directional error bias underscores a fundamental architectural limitation of Transformers: generative syntax dictates that constituent relationships are asymmetrical (a governing head c-commands its complement, but not vice versa), whereas softmax-normalized self-attention matrices calculate symmetric or unconstrained pairwise distributions. In deeply embedded structures (e.g., complex CS or nested DPs), intervening embedded nouns introduce competing agreement features. Without explicit structural biases enforcing directional dominance, self-attention heads remain prone to getting "distracted" by the matrix verb or primary nominal head (N1), leading to a systematic structural bias toward default high attachment. AraBERT was highly conservative in predicting N2 but comparatively more likely to select N1 when the structural evidence was insufficient to distinguish the embedded interpretation.

These results hold significant implications for the broader debate surrounding the cognitive and computational fidelity of LLMs. While the internal mechanisms of LLMs rely on statistical NNAs, not on formal generative grammars, our findings demonstrate that there is a promising role for neuro-symbolic "feeding" these NNAs and attention mechanisms with generative syntax notions, as LLMs cannot rely purely on surface data distribution to master complex syntax. The success of our candidate-based decision scheme ($G \in \{0, 1\}$) proves that injecting symbolic, rule-governed linguistic knowledge into Transformer pipelines yields superior results compared to end-to-end unconstrained learning (cf. Chatzikyriakidis & Lappin, 2026; Mahouachi, 2026).

Operationalizing a generatively informed neuro-symbolic framework for morphologically rich languages (MRLs) like Arabic enhances standard LLM finetuning forcing NNAs to jointly induce morphology, lexical semantics, agreement rules, and phrase structure simultaneously from raw surface text. This implication goes in line with Chatzikyriakidis and Lappin (2026) who provided some sort of evidence of possible integrating neuro-symbolic framework with NNSs (see also Gamage, 2025). In MSA, where rich inflectional agreement, clitics, null subjects, and CS ambiguities within complex DPs create extreme feature sparsity, unconstrained learning leads to erratic generalizations. Our framework demonstrates that injecting generative syntactic priors at the input-formatting level effectively bypasses the need for the model to "rediscover" constituent boundaries. By framing disambiguation as a candidate-based sequence-pair decision task ($S$, $c_{N1}$, $c_{N2}$), the model's self-attention parameters are freed from boundary detection and can focus entirely on evaluating contextual argument compatibility and long-range feature agreement.

Another related implication perhaps has perhaps to do with mitigating the asymmetrical "High-Attachment Bias" in LLM Attention Mechanisms. The pronounced recall asymmetry observed in complex Arabic DPs, where High/VP Attachment (N1) achieved 99.71% recall while Low/Embedded Attachment (N2) dropped to 89.26%, exposes a fundamental architectural limitation in current LLM working mechanisms. Standard $QK^T$ (or softmax-normalized attention) calculates unconstrained, pairwise token interactions that naturally favor surface proximity and matrix-level heads over embedded nominal domains. Recognizing that generative syntax dictates asymmetrical c-command relationships provides a clear roadmap for future LLM architectures (cf. Chomsky, 1981). Specifically, Arabic NLP models can be improved not just through parameter scaling (see also Liu et al., 2025), but by engineering syntactically constrained attention masks or c-command-aware relative positional encodings that prevent lower-embedded DP nodes (N2) from being absorbed into default higher-attachment (N1) interpretations.

A further implication based on our findings concerns scalable neuro-symbolic pipeline for prompt disambiguation in Arabic LLMs. In practical generative AI and LLM applications (such as Arabic machine translation, dialogue generation, and automated parsing), structural ambiguities in nominal phrases ([DP1 and DP2] + Modifier) often cause LLMs to hallucinate or select incorrect semantic interpretations (cf. Tzachristas & Sui, 2025). The candidate-based classification scheme ($G \in \{0, 1\}$) established in this study provides a reusable neuro-symbolic pipeline: a symbolic front-end where generative linguistic principles generate syntactically permissible candidate representations ($c_{N1}$, $c_{N2}$), and a neural back-end where the contextual Transformer (AraBERT/LLM) acts as a statistical discriminator to evaluate the contextual evidence for each candidate. This paradigm transforms opaque, unconstrained generation into an interpretable, candidate-guided selection process.

A final implication may perhaps be related to redefining the relationship between Chomskyan syntax and LLMs' working mechanism. The study provides evidence that viewing Chomskyan generative grammar and statistical NNAs as mutually exclusive paradigms needs serious rethinking and refinement, establishing a productive operational synthesis. Generative linguistics specifies the formal *search space* and structural hypotheses (what can structurally exist according to language-specific principles), while DL/NNA models quantify the probabilistic contextual evidence (what occurs given the surrounding discourse context). For Arabic computational linguistics, this hybrid methodology proves that formal syntactic theory is not to be replaced by scale, but an essential engineering tool for guiding, constraining, and interpreting modern DL architectures.

## 6. Conclusions and limitations

This study proposes a generatively informed neuro-symbolic framework for syntactic ambiguity resolution in MSA DPs, bridging formal generative syntax and deep contextual language modeling. By framing structural disambiguation as a candidate-based sequence pair decision task ($G \in \{0, 1\}$), we addressed a key limitation of standard Transformer architectures: their reliance on linear token proximity rather than directional, hierarchical phrase structure. Empirical evaluation of AraBERT on an unseen test set of 6696 human-annotated instances demonstrated exceptional performance, achieving an overall accuracy of 96.88%, a macro-F1 score of 95.92%, and a weighted F1 score of 96.83%. The findings confirm that providing Transformer LLMs with structured candidate priors constrains the search space during finetuning, allowing multi-head attention mechanisms to evaluate structural argument satisfaction and long-range morpho-syntactic agreement directly. Rather than forcing end-to-end models to induce complex phrase-structure trees unassisted, injecting formal syntactic assumptions proves to be an efficient, highly effective, and interpretable method for processing complex Arabic nominal, coordinate, and clausal constructions.

Despite the strong empirical results, several methodological and architectural limitations warrant discussion and offer avenues for future research: i) the recall gap manifests clearly between High/VP Attachment (N1: 99.71%) and Low/NP/Embedded Attachment (N2: 89.26%). While overall accuracy remains high, 195 N2 instances were misclassified as N1. This directional bias highlights the structural discrepancy between generative syntax's asymmetrical c-command relations and self-attention's symmetric matrix calculations. Future work should explore architectural modifications, such as custom tree-directed attention masks or c-command positional encodings, to enforce directional structural dominance directly within Transformer layers. ii) the current formulation operationalizes ambiguity as a binary choice between two explicitly generated candidate interpretations ($c_{N1}$ vs $c_{N2}$)). While this design cleanly isolates structural decision-making, natural language occasionally presents multi-way ambiguities involving three or more competing attachment sites (e.g., deeply stacked genitive CSs or multiple adverbial adjuncts). Extending the candidate-based framework to n-ary classification or ranking schemes remains an important direction for scaling this approach, and iii) our evaluation focused strictly on MSA corpus data with standardized preprocessing. Dialectal Arabic presents further syntactic variability, non-standard orthography, and reduced inflectional cues, which may increase structural ambiguity.

Additionally, expanding evaluation across completely unvocalized text versus partially diacritized text will clarify the exact extent to which AraBERT relies on explicit morphological markers versus syntactic context.

**Data availability**

The study notebook is available on GitHub here: https://github.com/Shormanim/A-generative-informed-neuro-symbolic-framework-for-syntactic-ambiguity-resolution